\documentclass[11pt]{article}

\PassOptionsToPackage{table}{xcolor}

\usepackage[preprint]{acl}

\usepackage{times}
\usepackage{latexsym}
\usepackage{amsmath}
\usepackage{amssymb}

\usepackage[T1]{fontenc}
\usepackage[utf8]{inputenc}

\usepackage{microtype}
\usepackage{inconsolata}

\usepackage{graphicx}
\usepackage{amsmath,amssymb,mathtools}
\usepackage{verbatim,enumitem,romannum}
\usepackage{multirow,booktabs,graphicx,makecell}
\usepackage{xcolor}
\usepackage{arydshln}
\usepackage{titletoc}
\usepackage{enumitem}

\newcommand{\vi}{\ensuremath{\mathcal{V\text{-}I}}}
\newcommand{\vv}{\ensuremath{\mathcal{V\text{-}V}}}
\newcommand{\method}{PRISM}
\newcommand{\clip}{CLIP}
\newcommand{\clipcite}{CLIP~\cite{radford2021learning}}

\newcommand{\sigliptwo}{SigLIP2}
\newcommand{\sigliptwocite}{SigLIP2~\cite{tschannen2025siglip}}

\newcommand{\timesformercite}{TimeSformer~\cite{bertasius2021space}}

\newcommand{\internvideocite}{InternVideo~\cite{wang2022internvideo}}

\newcommand{\egovlpcite}{EgoVLP~\cite{lin2022egocentric}}

\newcommand{\lavilacite}{\textsc{LaViLa}~\cite{zhao2023learning}}
\newcommand{\actorobservernet}{ActorObserverNet}
\newcommand{\actorobservernetcite}{ActorObserverNet~\cite{sigurdsson2018actor}}

\newcommand{\viencodercite}{VI Encoder~\cite{grauman2024ego}}

\newcommand{\egoinstructorcite}{EgoInstructor~\cite{xu2024retrieval}}
\newcommand{\suml}{SUM-L}
\newcommand{\sumlcite}{SUM-L~\cite{wang2023learning}}
\newcommand{\viewpointrosetta}{\textsc{ViewpointRosetta}}
\newcommand{\viewpointrosettacite}{\textsc{ViewpointRosetta}~\cite{luo2025viewpoint}}

\newcommand{\tcncite}{TCN~\cite{sermanet2018time}}

\newcommand{\carlcite}{CARL~\cite{chen2022frame}}

\newcommand{\tcccite}{TCC~\cite{dwibedi2019temporal}}

\newcommand{\gtacite}{GTA~\cite{hadji2021representation}}
\newcommand{\aetwo}{AE2}
\newcommand{\aetwocite}{AE2~\cite{xue2023learning}}

\newcommand{\vjepatwo}{V-JEPA~2}

\newcommand{\dinovtwo}{DINOv2}

\newcommand{\prismfull}{%
  \underline{P}re\-dic\-tive \underline{R}e\-com\-po\-si\-tion
  v\underline{I}a \underline{S}e\-man\-tic Latent Deco\underline{M}po\-si\-tion}

\title{PRISM: Predictive Recomposition via Semantic Latent Decomposition for View-invariant Video Representation Learning}

\author{
  Youngchae Chee\thanks{\ Equal contribution.} \\
  KAIST \\
  \small{\texttt{litcoderr@kaist.ac.kr}} \\\And
  Hosu Lee\footnotemark[1] \\
  KAIST \\
  \small{\texttt{leehosu01@kaist.ac.kr}} \\\And
  Sungjune Park\footnotemark[3] \\
  KAIST \\
  \small{\texttt{sungjune-p@kaist.ac.kr}} \\\AND
  Junho Kim\footnotemark[2] \\
  University of Illinois Urbana-Champaign \\
  \small{\texttt{arkimjh@illinois.edu}} \\\And
  Yong Man Ro\footnotemark[2] \\
  KAIST \\
  \small{\texttt{ymro@kaist.ac.kr}} \\
}

\begin{document}
\pagenumbering{arabic}
\maketitle
\footnotetext[2]{Corresponding authors.}
\footnotetext[3]{Currently working at LG AI Research.}
\begin{abstract}
%Learning video representations that remain consistent across egocentric and exocentric viewpoints is fundamental for understanding human behavior from any perspective, and language has emerged as a natural oracle for such view-invariant guidance. However, existing methods encode each video as a unified embedding, where view-invariant and view-variant semantics are prone to entanglement, undermining this very guidance. To resolve this, we propose PRISM, which leverages language compositionality that enforces disentanglement of the two factors into orthogonal representations. By further internalizing temporal dynamics into each stream, PRISM achieves state-of-the-art results on EgoExo4D, EgoExoLearn, AE2, and UNSCENE, including under zero-shot transfer.
Cross-view video representation learning aims to capture viewpoint-invariant action semantics despite substantial appearance changes across egocentric and exocentric videos. However, existing methods encode each video as a unified embedding, where view-invariant and view-variant semantics inevitably entangle under co-occurrences---a failure mode we show persists even in cross-view methods explicitly trained for view-invariance. Our key insight is that a view-invariant feature is truly disentangled when it can be sufficiently recomposed with an arbitrary view-variant feature while preserving their independent semantics. Building on this, we propose PRISM, that decomposes video into view-invariant and view-variant latents and recompose them under language supervision encouraging clean decomposition of the two streams. PRISM achieves state-of-the-art results on EgoExo4D, EgoExoLearn, AE2, even surpassing in-domain models under zero-shot setting. Code is available at \url{https://github.com/litcoderr/prism}.
\end{abstract}

\section{Introduction}

Human vision can naturally recognize actions across large viewpoint changes: although the same action may look substantially different when observed from a first-person (\textit{ego}) or third-person (\textit{exo}) viewpoint, it is often perceived as semantically identical~\cite{isik2018fast}. This property enables higher-order skill acquisition---for example, a learner can observe an expert performing a skill from an exo viewpoint and later refine their own attempt by semantically aligning their ego experience with the previously observed demonstration. 

In this context, view-invariant representation learning~\cite{sigurdsson2018actor, ardeshir2018exocentric, grauman2024ego, sermanet2018time} has emerged as a core challenge for generalizing action semantics. It underpins a broad spectrum of applications, ranging from cross-view action recognition and video understanding~\citep{huang2024egoexolearn, xue2023learning, sener2022assembly101, sigurdsson2018charades} to robotics~\citep{sermanet2018time, pang2025learning}.

\begin{figure}[t]
    \centering
    \includegraphics[width=\linewidth]{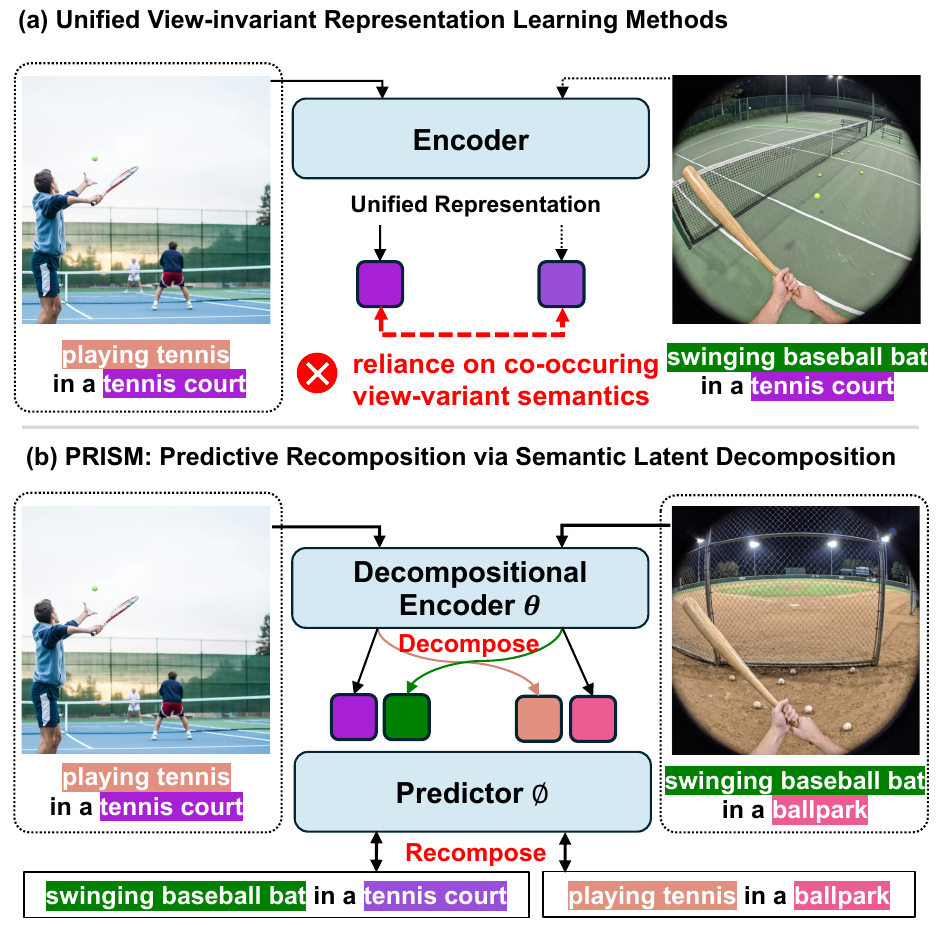}
    \vspace{-7mm}
    \caption{\textbf{Comparison of unified and compositional representations.} (a)  Unified representations entangle  actions (\textit{playing tennis}) with co-occurring context (\textit{tennis court}), failing under novel compositions. (b) PRISM decomposes video into view-invariant and view-variant latents and recomposes them under language supervision, enforcing clean disentanglement.}
    \label{fig:teaser}
    \vspace{-5mm}
\end{figure}

% Early efforts~\citep{ardeshir2018exocentric, grauman2024ego, sermanet2018time, sigurdsson2018actor} have relied on multi-view datasets that are time-synchronized across views. To remove the temporal dependency on the dataset, more recent works extend the problem to unpaired setup that can be collected at scale~\citep{xue2023learning, huang2024egoexolearn}.
% Some extension that focuses on fine-grained temporal modeling have relied on heuristics such as ROI pooling with hand-object detector~\citep{xue2023learning} or selecting the top-$K$ patches whose features change the most between adjacent frames~\citep{park2025bootstrap}. Nonetheless, such heuristic schemes cannot guarantee strict view invariance.
% Another line of work leverages language---either to construct ego--exo pseudo-pairs~\citep{wang2023learning, luo2025viewpoint} or to align video features directly with a view-invariant language description embedding space~\citep{xu2024retrieval}---to induce view invariance.

Early efforts learn a unified view-invariant representation by aligning visual features temporally from two different views~\citep{sermanet2018time, xue2023learning}, but they usually lack semantic disentanglement.
To overcome this, subsequent works leverage language as a view-invariant semantic signal, either by constructing ego–exo pseudo-pairs based on description similarities~\citep{wang2023learning} or by directly aligning video features with the descriptions in a language embedding space~\citep{xu2024retrieval, luo2025viewpoint}.

% However, previous methods that encode a video into a single unified representation have failed to prevent view-variant information from leaking in when a dataset exhibits a strong correlation between view-invariant(\vi) linguistic descriptions and the underlying view-variant(\vv) content.
% As we illustrate in Figure~\ref{fig:teaser}, if the \vi\ action ``playing tennis'' co-occurs predominantly with the \vv\ background ``tennis court'', a model trained on such data is prone to falsely associating ``playing tennis'' related clips with clips of different action but same ``tennis court'' \vv\ semantics---a critical failure mode.
% 
% We therefore aim to learn a representation that cleanly decompose \vi\ features from \vv\ features while still capturing fine-grained temporal dynamics. Our key insight is the following: given a video, its \vi\ context (\textit{e.g.,} playing tennis) is perfectly decomposed if and only if the model can still predict the semantic composition of this \vi\ context with an off-distribution \vv\ context (\textit{e.g., garage}).
% Thus we propose \textbf{PRISM}: \underline{P}redictive \underline{R}ecomposition v\underline{I}a \underline{S}emantic Latent Deco\underline{M}position, a framework that learns orthogonally decomposed view-invariant representations leveraging language.

Despite these advances, previous works are bounded by the same limitation: they collapse a video into a single unified representation, which fails to prevent view-variant information from leaking in when a dataset exhibits a strong correlation between view-invariant (\vi) semantics and the underlying view-variant (\vv) content. As illustrated in Fig.~\ref{fig:teaser}, if the \vi\ action ``\textit{playing tennis}'' predominantly co-occurs with the \vv\ context ``\textit{tennis court}’’, unified representations may rely on the shared contextual semantics rather than the action identity itself, causing \vi\ and \vv\ semantics to become entangled.

We argue that a truly \vi\ representation must be independently decomposed from \vv\ semantics. Our key insight is that such decomposition is only achieved when a model can freely recompose a \vi\ representation with an off-distribution \vv\ counterpart while still preserving the correct semantic meaning. Thanks to its inherently compositional structure, language naturally provides semantically controllable \vi\ / \vv\ compositions, enabling seamless decomposition and novel semantic recombination beyond observed visual co-occurrences. Moreover, language captures high-level conceptual identity independently of the low-level visual variations induced by viewpoint changes.

In this paper, we introduce \textbf{PRISM}: \prismfull, a framework that learns semantically decomposed view-invariant representations through language supervision. PRISM enforces an encoder to decompose an input video into \vi\ and \vv\ representations such that, when recombined with off-distribution counterparts, a predictor can correctly reconstruct the corresponding semantic composition in language latent space. While language provides strong supervision for clip-level semantic alignment, it lacks fine-grained temporal specificity. To address this limitation, we further introduce a frame-level self-predictive objective that internalizes temporal dynamics alongside clip-level semantics.

%Through extensive evaluations on cross-view video understanding benchmarks~\citep{huang2024egoexolearn, luo2025viewpoint, xue2023learning}, we show that \textbf{PRISM} achieves superior view-invariant representation as well as effectively capturing fine-grained temporal dynamics--surpassing state-of-the-art methods by a big margin even in a zero-shot setting. Furthermore, we validate the robustness to background correlation on ~\citep{bae2024devias} and conduct rigorous analysis.
Through extensive evaluations on cross-view video understanding benchmarks~\citep{huang2024egoexolearn, luo2025viewpoint, xue2023learning}, we corroborate that PRISM learns superior view-invariant representations while effectively capturing fine-grained temporal dynamics, outperforming state-of-the-art methods even in zero-shot settings and surpassing in-domain models on temporal phase prediction. Furthermore, on UNSCENE~\citep{bae2025mash}, PRISM substantially improves retrieval performance under background correlation shifts compared to prior cross-view methods.

% Contributions 1, 2, 3
In summary, our main contributions are:
\begin{itemize}[leftmargin=*]
    \item{We identify a critical failure mode where unified view-invariant representation learning methods entangle view-variant semantics under biased co-occurrences.}
    \item{We propose \textbf{PRISM}: \prismfull, which leverages language compositionality to semantically decompose view-invariant and view-variant semantics, achieving state of the art performance in cross-view video understanding benchmarks.}
    \item{We devise a frame-level self-predictive objective that complements language supervision, internalizing fine-grained temporal dynamics without compromising semantic disentanglement.}
    %\item{We leverage language as \textbf{PRISM}: \underline{P}redictive \underline{R}ecomposition v\underline{I}a \underline{S}emantic Latent Deco\underline{M}position, }
    %\item{We propose \textbf{PRISM}: \underline{P}redictive \underline{R}ecomposition v\underline{I}a \underline{S}emantic Latent Deco\underline{M}position, a simple yet highly effective framework that encourages clean decomposition of view-invariant semantics leveraging language.}
    % \item{We demonstrate that by integrating clip-level language supervision with a frame-level self-predictive objective, PRISM successfully captures fine-grained temporal dynamics without compromising semantic disentanglement.}
    % \item{We empirically validate that our factorized representation effectively mitigates spurious correlations between view-invariant and view-variant semantics, demonstrating robustness on ** unusual context videos compared to existing unified representation view-invariance methods.}
\end{itemize}
\section{Related Work}

\subsection{View-Invariant Representation Learning}

With the rise of first-person interactive platforms such as augmented reality and robotics, visual recognition from the egocentric viewpoint has become a critical challenge. Large-scale vision-language models (\textit{e.g.,} CLIP~\cite{radford2021learning}, SigLIP2~\cite{tschannen2025siglip}) provide powerful visual representations through vision-language alignment, yet remain tied to the training viewpoint and struggle to maintain semantic consistency across egocentric and exocentric perspectives.
To address this limitation, view-invariant representation learning aims to produce consistent features regardless of camera viewpoint. Early efforts relied on time-synchronized multi-view datasets~\cite{sigurdsson2018charades, grauman2024ego}, while more recent works extend to unpaired ego-exo videos collectible at scale~\cite{huang2024egoexolearn, xue2023learning}. SUM-L~\cite{wang2023learning} aligns unpaired videos through language-based semantic matching, and AE2~\cite{xue2023learning} introduces a fine-grained cross-view benchmark with temporal alignment objectives. More recently, ViewpointRosetta~\cite{luo2025viewpoint} proposes a diffusion-based translator that synthesizes exo features from ego features jointly align hallucinated cross-view feature along with language.

\begin{figure*}[t]
    \centering
    \includegraphics[width=\linewidth]{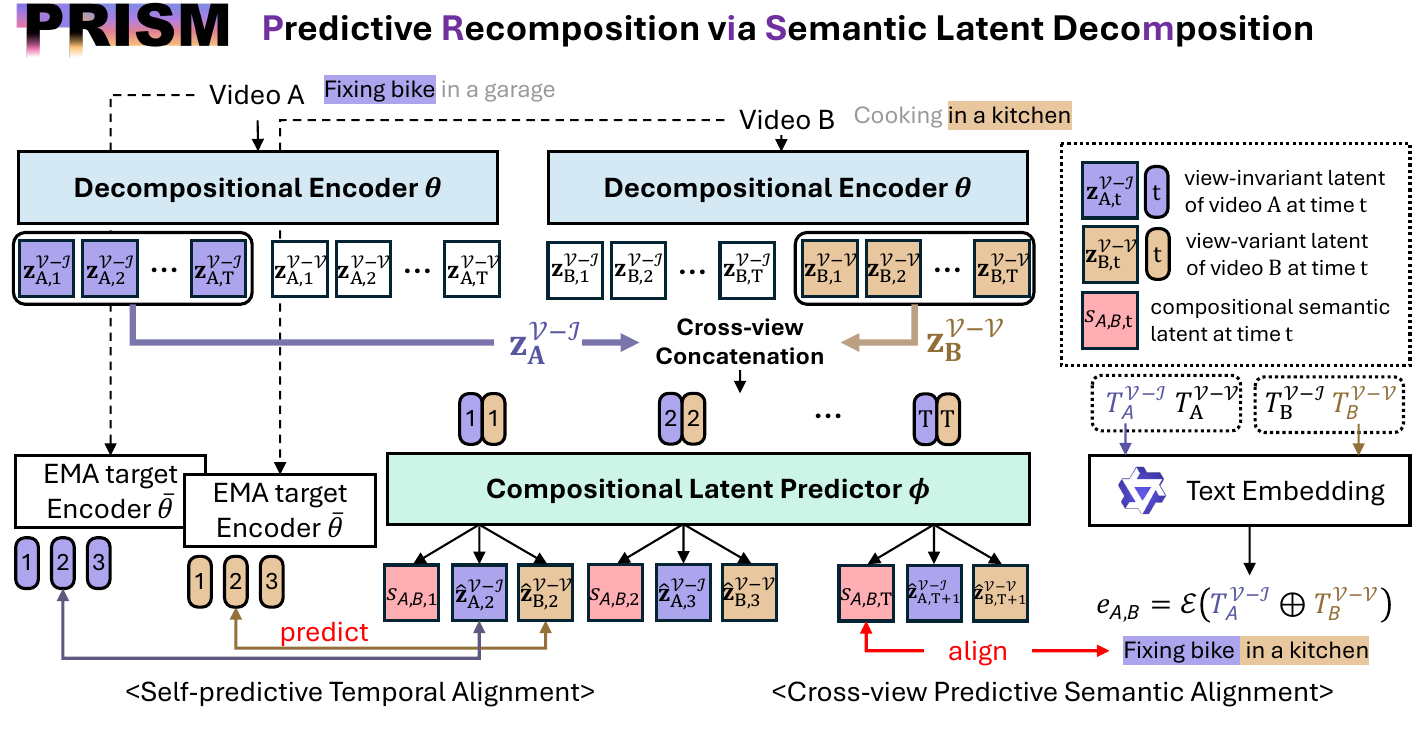}
    \vspace{-8mm}
    \caption{\textbf{Overview of PRISM.} PRISM decomposes videos into view-invariant (\vi) and view-variant (\vv) latent streams, cross-composes them across videos, and aligns the resulting compositional latent with recombined language semantics. A self-predictive temporal objective further internalizes fine-grained temporal dynamics.}
    \label{fig:method}
\end{figure*}
\subsection{Action-Scene Entanglement}

The spurious correlation between actions and their background scenes is a persistent bias in video understanding, as evidenced by RESOUND~\cite{li2018resound}, which show models often exploit scene cues rather than motion semantics. DEVIAS~\cite{bae2024devias} further reveals that such models suffer severe performance degradation under unseen action-scene compositions, while MASH-VLM~\cite{bae2025mash} identifies similar action-scene hallucinations in Video-LLMs and proposes disentangled attention to mitigate them.
However, existing multi-view alignment methods~\cite{xu2024retrieval, wang2023learning, luo2025viewpoint} indiscriminately minimize the distance between co-observed features, failing to distinguish whether the extracted commonality originates from the view-invariant action or the view-variant background.

\vspace{3mm}
\section{Proposed Method}
% The method decomposes video features into view-invariant and view-variant components, training via cross-composition alignment with language semantics. This enforces orthogonal, disentangled representation learning.
\paragraph{\method\ Overview.} As shown in Fig.~\ref{fig:method}, the core principle of \method\ is that visual information in a video can be decomposed into a view-invariant component \vi\ and a view-variant component \vv. If this decomposition is achieved cleanly, the \vi\ of one video should retain its semantic identity even when recomposed with the \vv\ of another video. Building on this, we enforce that a visual representation formed by cross-composing the decomposed \vi\ of video \textsf{A} with the \vv\ of video \textsf{B} aligns with the corresponding recomposed semantics at the language level (and vice versa for the reverse composition). By structurally preventing one factor from leaking into the other, this objective encourages the encoder to learn orthogonally decomposed representations without entanglement. We use \textit{orthogonal} throughout in this semantic sense: the two streams are required to carry the disjoint language-level semantics of \vi\ and \vv, rather than to be geometrically orthogonal vectors.
% As illustrated in Fig.~\ref{fig:overview}, the core principle of \method\ is to decompose visual information in a video into a view-invariant component \vi\ and a view-variant component \vv. The core idea is that a visual representation formed by recomposing the decomposed \vi\ and \vv\ of two distinct videos, $\mathsf{A}$ and $\mathsf{B}$, must perfectly align with their semantically recombined meaning at the language level.
% 추가된 코멘트 볼것 넵 굳

To realize this, our framework comprises two modules: (\lowercase\expandafter{\romannumeral1}) Decompositional Encoder $\theta$ that decomposes a video into \vi\ and \vv\ representations, and (\lowercase\expandafter{\romannumeral2}) Compositional Latent Predictor $\phi$ that cross-composes representations from two distinct videos and synthesizes a compositional semantic embedding. This embedding is supervised to align with a sentence embedding constructed by combining decoupled language descriptions generated via a pre-trained LVLM~\cite{bai2025qwen3, gemini3flash} (\S\ref{sec:compose-decompose}, \S\ref{sec:method-decompose}). Finally, to overcome the limited temporal resolution of clip-level linguistic supervision, we introduce a self-predictive signal that tasks $\phi$ with predicting the future representations of $\theta$, enabling the model to internalize fine-grained temporal dynamics alongside semantic disentanglement (\S\ref{sec:method-temporal}).
% \paragraph{\method\ Overview.}
% Our framework consists of two modules: (\lowercase\expandafter{\romannumeral1}) a Decompositional Encoder $\theta$ that takes a video and decomposes it into two frame-level representations, $z^\vi$ and $z^\vv$, and (\lowercase\expandafter{\romannumeral2}) a Compositional Latent Predictor $\phi$ that takes the representations $z_\mathsf{A}^\vi$ and $z_\mathsf{B}^\vv$ from two independently sampled videos to recompose a compositional semantic embedding $s_{\mathsf{A}, \mathsf{B}}$ (and vice versa for $s_{\mathsf{B}, \mathsf{A}}$, omitted for simplicity). This recomposed representation $s_{\mathsf{A}, \mathsf{B}}$ is supervised to align with a sentence embedding constructed by combining decoupled language-level descriptions, $T_\mathsf{A}^\vi$ and $T_\mathsf{B}^\vv$, which are generated via a pre-trained LVLM. By cross-composing elements from independent videos, we ensure that the counterpart factors (\textit{e.g.,} $z_\mathsf{A}^\vv$ or $T_\mathsf{A}^\vv$) cannot influence the predictor's output $s_{\mathsf{A}, \mathsf{B}}$. This structurally prevents information leakage, encouraging $\theta$ to learn a clean decomposition without entanglement. 
% % 여기 밑에 pargarph는 여기에 안넣어도 될듯한데?
% % Finally, to overcome the limited temporal resolution of clip-level linguistic supervision, we introduce a self-predictive signal that tasks the predictor $\phi$ with predicting the future representations of the encoder $\theta$, allowing the model to internalize fine-grained temporal dynamics.

\subsection{Decompose-and-Recompose Schema} \label{sec:compose-decompose}
% \subsection{Decomposion-and-Recomposition} \label{sec:compose-decompose}

To operationalize our core principle, we first define the structural mechanism that handles the visual features before introducing any external supervision. We feed an input video $v$ into our Decompositional Encoder $\theta$ to separate the primary interaction \vi\ from the view-variant context \vv, producing two frame-level representations:
\begin{equation}
    (\mathbf{z}^\vi_{v},\ \mathbf{z}^\vv_{v}) = \theta(v).
\end{equation}
Once decomposed, we deliberately break each video’s natural co-occurrence through cross-composition. Given two independently sampled videos $\mathsf{A}$ and $\mathsf{B}$, we perform \textit{cross-view concatenation} by pairing $\mathbf{z}^\vi_\mathsf{A}$ with $\mathbf{z}^\vv_\mathsf{B}$ (and vice versa for the reverse composition, omitted for simplicity). This cross-composed pair is fed into the Compositional Latent Predictor $\phi$ to synthesize a compositional semantic latent:
\begin{equation}
    s_{\mathsf{A},\mathsf{B}} = \phi(\mathbf{z}^\vi_{\mathsf{A}},\ \mathbf{z}^\vv_{\mathsf{B}}).
\end{equation}
By forcing the representations through this \textit{decompose and recompose} bottleneck, $\phi$ must construct a coherent semantic embedding without access to the original, intact video composition.

% With the recomposed latent $s_{\mathsf{A},\mathsf{B}}$ (or $s_{\mathsf{B},\mathsf{A}}$) established, our next step builds upon standard vision-language contrastive modeling~\citep{}: beyond merely aligning holistic visual and linguistic representations, we enforce alignment between cross-composed visual latents and their correspondingly recombined language descriptions. 

% We feed an input video $v$ into our Decompositional Encoder $\theta$ to separate the primary interaction \vi\ from the background context \vv, producing two frame-level representations:
% \begin{equation}
%     (\mathbf{z}^\vi_{v},\ \mathbf{z}^\vv_{v}) = \theta(v).
% \end{equation}
% Given two independently sampled videos $\mathsf{A}$ and $\mathsf{B}$, we perform \textit{cross-view concatenation} by pairing $\mathbf{z}^\vi_\mathsf{A}$ with $\mathbf{z}^\vv_\mathsf{B}$, deliberately breaking each video's natural co-occurrence. This cross-composed pair is fed into the Compositional Latent Predictor $\phi$ to synthesize a compositional semantic latent:
% % Given two independently sampled videos $\mathsf{A}$ and $\mathsf{B}$, we take the decomposed representations from each and feed them into the Compositional Latent Predictor $\phi$ to synthesize a compositional semantic latent:
% \begin{equation}
%     s_{\mathsf{A},\mathsf{B}} = \phi(\mathbf{z}^\vi_{\mathsf{A}},\ \mathbf{z}^\vv_{\mathsf{B}}).
% \end{equation}

\subsection{Language Supervised Decomposition} \label{sec:method-decompose}

With the recomposed latent $s_{\mathsf{A},\mathsf{B}}$ (or $s_{\mathsf{B},\mathsf{A}}$), we next enforce that cross-composed visual representations preserve the correct recomposed semantics at the language level. The key idea is that spurious correlations between \vi\ and \vv\ semantics become exposed under cross-composition, since the original co-occurrence structure is intentionally broken. As a result, a model that relies on shortcut dependencies between the two factors will fail to reconstruct the correct recomposed semantics.

To supervise the correct recomposed semantics, we construct a target representation at the language level. Specifically, we combine the view-invariant description $T^\vi_{\mathsf{A}}$ of video $\mathsf{A}$ with the view-variant description $T^\vv_{\mathsf{B}}$ of video $\mathsf{B}$, both generated through a pre-trained LVLM~\cite{bai2025qwen3, gemini3flash}. Intuitively, if $\phi$ relies on shortcut correlations in $\mathbf{z}^\vv$ to infer the semantics of $\mathbf{z}^\vi$ (\textit{e.g.,} kitchen $\rightarrow$ cooking), such dependencies become inconsistent under cross-composition and therefore cannot match the recomposed language target. A text embedding model~\cite{zhang2025qwen3} $\mathcal{E}$ then maps the recombined text into a target semantic embedding:
\begin{equation}
    e_{\mathsf{A},\mathsf{B}} = \mathcal{E}(T^\vi_{\mathsf{A}} \oplus T^\vv_{\mathsf{B}}).
\end{equation}

We then apply a contrastive objective that aligns the compositional latent $s_{\mathsf{A},\mathsf{B}}$ with the recombined text embedding $e_{\mathsf{A},\mathsf{B}}$. With a temperature parameter $\tau$ and a similarity kernel $\mathbb{K}(\mathbf{x}, \mathbf{y}) = \exp(\mathrm{sim}(\mathbf{x}, \mathbf{y})/\tau)$, we define the decomposition loss $\mathcal{L}_\text{decomp}$ as:
\begin{equation}
    \mathcal{L}_\text{decomp} = \mathbb{E}_{\mathsf{A},\mathsf{B} \in \mathcal{B}} \left[ - \log \frac{\mathbb{K}(s_{\mathsf{A},\mathsf{B}}, e_{\mathsf{A},\mathsf{B}})}{\sum_{\mathsf{C},\mathsf{D} \in \mathcal{B}} \mathbb{K}(s_{\mathsf{A},\mathsf{B}}, e_{\mathsf{C},\mathsf{D}})} \right].
\end{equation}

\subsection{Internalizing Temporal Dynamics} \label{sec:method-temporal}

While the language-based objective $\mathcal{L}_\text{decomp}$ successfully achieves semantic decomposition of visual information, clip-level linguistic supervision only provides coarse semantic summaries, such as ``\textit{chopping an onion}''. Consequently, it is insufficient for learning representations capable of capturing fine-grained frame-level temporal dynamics, such as ``\textit{lifting a knife} $\rightarrow$ \textit{placing it on the onion} $\rightarrow$ \textit{slicing downward}''. To compensate for this limited temporal resolution and internalize fine-grained temporal structures into the learned representations, we introduce a self-predictive signal that tasks the predictor $\phi$ with forecasting the future outputs of the encoder $\theta$.

Concretely, given the accumulated representations $\mathbf{z}^\vi_{\le t}$ and $\mathbf{z}^\vv_{\le t}$ up to time $t$, $\phi$ generates predictions for the next-frame representations:
\begin{equation}
    (\hat{\mathbf{z}}^\vi_{t+1},\ \hat{\mathbf{z}}^\vv_{t+1}) = \phi(\mathbf{z}^\vi_{\le t},\ \mathbf{z}^\vv_{\le t}).
\end{equation}
The prediction targets for this objective are the encoder $\theta$'s own future outputs. Since $\theta$ is continuously updated during training, using $\theta$ directly as the target encoder would yield unstable supervisory signals and risk representational collapse. To provide stable learning targets, we utilize a target encoder $\bar{\theta}$, parameterized by an Exponential Moving Average (EMA) of $\theta$'s weights. At each training step, the target weights are updated as follows, after which the target frame-level representations are extracted from a given video $v$:
\begin{equation}
    \bar{\theta} \leftarrow \alpha \bar{\theta} + (1 - \alpha) \theta,
\end{equation}
\begin{equation}
    (\bar{\mathbf{z}}^{\vi}_{v},\ \bar{\mathbf{z}}^{\vv}_{v}) = \bar{\theta}(v).
\end{equation}
% Finally, we define an objective function that maximizes the cosine similarity between the predicted and target representations. Through the following temporal objective $\mathcal{L}_\text{temp}$, each stream (\vi\ and \vv) independently forms its own temporal dynamics without relying on external alignment signals:
Finally, we define the temporal objective $\mathcal{L}_\text{temp}$, which maximizes the cosine similarity between predicted and target representations, allowing each stream (\vi\ and \vv) to capture its own temporal dynamics independently without relying on external alignment signals:
\begin{equation}
    \mathcal{L}_\text{temp} = \mathbb{E}_{c \in \{\vi, \vv\}} \left[ - \text{sim}(\hat{\mathbf{z}}^c_{t}, \bar{\mathbf{z}}^c_{t}) \right].
\end{equation}
This objective incentivizes the encoder $\theta$ to embed temporal foresight into its representation space. Consequently, rather than merely capturing the present state, the representations at each time step naturally anticipate upcoming transitions.

\definecolor{headerblue}{HTML}{ECF4FF}
\definecolor{headergray}{HTML}{F2F2F2}
\begin{table*}[t]
\centering
\small
\setlength{\tabcolsep}{1pt}
% \caption{Cross-view evaluation results.}
% \label{tab:crossview_bench}
\vspace{-2mm}
\resizebox{1.0\linewidth}{!}{
    \begin{tabular}{@{}l ccc cc c ccc ccccc c @{}}
\toprule
% \multirow{3}{*}[-4pt]{Method}
% & \multicolumn{5}{c}{\textbf{Paired-View}}
% & \multicolumn{7}{c}{\textbf{Unpaired-View}} \\
\multirow{3}{*}[-4pt]{Method}
& \multicolumn{6}{c}{\textbf{EgoExo4D}}
& \multicolumn{9}{c}{\textbf{EgoExoLearn}} \\
\cmidrule(l{2pt}r{2pt}){2-7} \cmidrule(l{2pt}r{2pt}){8-16}
& \multicolumn{3}{c}{Retrieval (R@5)}
& \multicolumn{2}{c}{Recognition}
& Assessment
& \multicolumn{3}{c}{Association$_\text{test}$}
& \multicolumn{5}{c}{Anticipation (R@5)}
& Assessment \\
\cmidrule(l{2pt}r{2pt}){2-4} \cmidrule(l{2pt}r{2pt}){5-6} \cmidrule(l{2pt}r{2pt}){7-7} \cmidrule(l{2pt}r{2pt}){8-10} \cmidrule(l{2pt}r{2pt}){11-15} \cmidrule(l{2pt}r{2pt}){16-16}
& ego2exo & exo2ego & avg & top-1 & top-5 & Acc & ego2exo & exo2ego & avg & ego-V & ego-N & exo-V & exo-N & avg & Acc \\
\hline
\rowcolor{headergray} \multicolumn{16}{l}{\textit{Image-Language Model}} \\
\clipcite              & 19.11 & 12.24 & 15.68 & 10.49 & 29.90 & 54.93 & 16.00 & 15.64 & 15.82 & 33.50 & 37.40 & 39.60 & 44.30 & 38.70 & 73.48 \\
\sigliptwocite         & 35.08 & 19.72 & 27.40 & 13.86 & 37.84 & \underline{55.57} & 25.00 & 28.10 & 26.6 & 64.70 & 71.70 & \underline{56.90} & 65.00 & 64.60 & \textbf{76.03} \\
\hline
\rowcolor{headergray} \multicolumn{16}{l}{\textit{Video-Language Model}} \\
\timesformercite       & 6.68 & 6.95 & 6.82 & 5.18 & 14.14 & 51.58 & 15.00 & 17.64  & 16.32 & 68.09 & 75.18 & 51.57 & 61.92 & 64.19 & \underline{75.47} \\
\internvideocite       & 23.64 & 20.67 & 22.16 & 13.67 & 38.99 & 52.58 & 30.60 & 21.70 & 26.20 & 63.76 & 74.28 & 56.67 & \textbf{65.92} & \underline{65.16} & 68.88 \\
\hline
\rowcolor{headerblue} \multicolumn{16}{l}{\textit{Ego-Centric Methods}} \\
\egovlpcite            & 29.33 & 13.47 & 21.40 & 20.33 & 46.32 & 54.37 & 31.50 & 28.90 & 30.20 & 67.70 & 72.50 & 50.40 & 52.00 & 60.65 & 69.04\\
\lavilacite            & 34.91 & 12.02 & 23.47 & 26.43 & 55.01 & 54.10 & 28.70 & 25.70 & 27.20 & \textbf{75.67} & \underline{76.58} & 48.03 & 51.05 & 62.83 & 68.44 \\
\hline
\rowcolor{headerblue} \multicolumn{16}{l}{\textit{Cross-View Methods}} \\
\actorobservernetcite  & 29.50 & 24.85 & 27.18 & 15.70 & 38.45 & 54.10 & 11.91 & 11.36 & 11.64 & 63.50 & 61.60 & 49.10 & 50.30 & 56.13 & 68.59 \\
\viencodercite         & 29.53 & 24.40 & 26.97 & 14.85 & 35.39 & 53.83 & - & - & - & - & - & - & - & - & - \\
% \egoinstructor     & 46.04 & 31.68 & 38.86 & 24.15 & 51.4 & 54.73 & 22.09 & 17.82 & 19.96 & 75.96 & 77.66 & 56.77 & 62.67 & 68.27 & 70.93 \\
\egoinstructorcite     & 46.04 & 31.68 & 38.86 & 24.15 & 51.40 & 54.73 & - & - & - & - & - & - & - & - & -   \\
\sumlcite              & 47.14 & 32.77 & 39.96 & 24.83 & 52.08 & 55.10 & 5.18 & 4.09 & 4.64 & 61.4 & 61.70 & 34.70 & 24.20 & 45.50 & 65.31 \\
\viewpointrosettacite  & \underline{58.14} & \underline{47.21} & \underline{52.68} & \underline{34.47} & \underline{64.85} & \textbf{55.82} & \underline{33.36} & \underline{31.27} & \underline{32.32} & 66.44 & 72.21 & 52.82 & 57.10 & 62.14 & 73.70 \\
\cdashline{1-16}\noalign{\vskip 0.5ex}
\method            & \textbf{75.89} & \textbf{50.27} & \textbf{63.08} & \textbf{41.93} & \textbf{72.92} & 55.28 & \textbf{44.36} & \textbf{43.36} & \textbf{43.86} & \underline{72.78} & \textbf{77.15} & \textbf{62.58} & \underline{65.33} & \textbf{69.46} & 68.53 \\
\bottomrule
\end{tabular}
}
\vspace{-1mm}
\caption{\textbf{Cross-view semantic alignment on EgoExo4D and EgoExoLearn.} We compare VLMs, ego-centric and cross-view methods. \method\ achieves new SoTA on Retrieval, Recognition, Association, and Anticipation.}
\label{tab:crossview_bench}
\vspace{-1mm}
\end{table*}
% \footnotetext[1]{Evaluated with linear probing.}
\section{Experiments}
We evaluate the capability of \method\ to comprehend video semantics and identify activities \S\ref{sec:semantic-align}, its ability to model fine-grained temporal stages and sequences \S\ref{sec:temporal-modeling}, and its robustness against performance degradation in scenarios where the activity is highly disentangled from the background context \S\ref{sec:background-robustness}. Furthermore, we present ablation studies to validate the impact of individual core components on overall performance \S\ref{sec:ablations}, concluding with an analysis of the focal points within the fully trained \method's latent representations \S\ref{sec:analysis}.

\definecolor{headerblue}{HTML}{ECF4FF}
\definecolor{headergray}{HTML}{F2F2F2}
\begin{table*}[t]
\centering
\small
\setlength{\tabcolsep}{2pt}
% \caption{zero shot .. }
% \label{tab:phase_bench}
\vspace{-2mm}
\resizebox{1.0\linewidth}{!}{
    \begin{tabular}{@{}l cccc c @{\hspace{20pt}}cccc c @{}}
\toprule
\multirow{2}{*}[-4pt]{Method}
& \multicolumn{4}{c}{\textbf{Frame Retrieval (mAP@10)}}
& \multirow{2}{*}[-2pt]{\textbf{Kendall's $\tau$}}
& \multicolumn{4}{c}{\textbf{Action Phase Classification (F1)}}
& \multirow{2}{*}[-2pt]{\shortstack{\textbf{Phase} \\ \textbf{Progression}}} \\
\cmidrule(l{2pt}r{2pt}){2-5} \cmidrule(l{2pt}r{2pt}){7-10}
& regular & ego2exo & exo2ego & avg & & regular & ego2exo & exo2ego & avg & \\
\hline
Random           & 53.97 & 51.68 & 51.20 & 52.28 & 0.004 & 32.90 & 33.44 & 33.46 & 33.27 & $-$0.069 \\
\hline
% \rowcolor{headergray} \multicolumn{11}{l}{\textit{In-Domain Pre-training (Trained on \aetwo\ videos)}} \\
\rowcolor{headergray} \multicolumn{11}{l}{\textit{Trained w/ AE2 videos (In-Domain)}} \\
\actorobservernetcite & 50.47 & 42.70 & 41.29 & 44.82 & 0.002 & 36.14 & 36.40 & 31.00 & 34.51 & $-$0.052 \\
\tcncite              & 58.25 & 47.37 & 42.48 & 49.37 & 0.046 & 56.80 & 35.92 & 41.40 & 44.71 & $-$0.227 \\
\carlcite             & 56.44 & 51.14 & 47.86 & 51.81 & 0.025 & 52.22 & 40.85 & 43.19 & 45.42 & $-$0.124 \\
\tcccite              & 70.58 & 62.08 & 65.84 & 66.17 & 0.400 & 67.17 & 55.90 & 52.27 & 58.45 &    0.322 \\
\gtacite              & 72.42 & 66.39 & 65.45 & 68.08 & 0.464 & 69.63 & 63.29 & 70.41 & 67.77 &    0.322 \\
\aetwocite            & 75.78 & 72.58 & 71.25 & 73.20 & 0.562 & 75.96 & 71.00 & 76.44 & 74.47 &    0.480 \\
% \byov             & 79.56 & 80.44 & 78.26 & 79.42 & 0.746 & 82.34 & 81.60 & 79.96 & 81.30 &    0.747 \\
\hline
% \rowcolor{headerblue} \multicolumn{11}{l}{\textit{Out-of-Domain Pre-training (Zero-Shot Transfer / No AE2 videos)}} \\
\rowcolor{headerblue} \multicolumn{11}{l}{\textit{Trained w/o AE2 videos (Zero-Shot Transfer)}} \\
ResNet-50~\cite{he2016deep}        & 58.26 & 44.87 & 42.94 & 48.69 & 0.025 & 53.53 & 31.47 & 45.25 & 43.41 & $-$1.215 \\
\clipcite            & 53.70 & 47.35 & 41.27 & 47.44 & 0.047 & 55.83 & 39.39 & 37.46 & 44.23 & $-$1.212 \\
\sigliptwocite       & 49.87 & 45.58 & 41.23 & 45.56 & 0.020 & 52.39 & 38.00 & 41.33 & 43.91 & $-$1.322 \\
%\actorobservernet& &  &  &  &  &  &  &  &  & \\
% \viencoder       & &  &  &  &  &  &  &  &  & \\
\sumlcite            & 56.20 & 39.70 & 41.50 & 45.80 & 0.110 & 63.00 & 43.80 & 46.00 & 50.93 &    0.030 \\
\viewpointrosettacite & 60.80 & 53.50 & 48.20 & 54.17 & 0.047 & 53.20 & 39.00 & 48.60 & 46.93 & $-$0.150 \\
\cdashline{1-11}\noalign{\vskip 0.5ex}
%\method            & 76.34 & 68.44 & 63.19 & 69.32 & 0.773 & 83.25 & 79.10 & 75.21 & 79.19 &    0.721 \\
% \method           & \textbf{76.03} & \textbf{70.05} & \textbf{70.51} & \textbf{72.20} & \textbf{0.863} & \textbf{83.05} & \textbf{77.98} & \textbf{77.68} & \textbf{79.57} & \textbf{0.738} \\
\method           & \textbf{77.70} & \textbf{68.90} & \textbf{65.00} & \textbf{70.53} & \textbf{0.601} & \textbf{79.60} & \textbf{69.80} & \textbf{71.30} & \textbf{73.57} & \textbf{0.647} \\
\bottomrule
\end{tabular}
}
\vspace{-1mm}
\caption{\textbf{Fine-grained temporal modeling on the AE2 benchmark.} We compare methods trained w/ and w/o AE2 videos. \method\ achieves new SoTA among methods trained w/o AE2 data.}
\label{tab:phase_bench}
\vspace{-1mm}
\end{table*}

\subsection{Cross-View Semantic Alignment}\label{sec:semantic-align}
\paragraph{Evaluation Setup.} We evaluate \method\ on the EgoExo4D~\cite{grauman2024ego} and EgoExoLearn~\cite{huang2024egoexolearn} benchmarks to assess its ability to correctly comprehend actions and recognize their intrinsic semantic equivalence. The specific evaluation tasks are defined as follows: \textit{Retrieval} measures the model's ability to identify different views (\textit{i.e.,} \textit{ego} and \textit{exo}) of the same RoI as semantically identical. \textit{Recognition} is the task of classifying the specific action occurring within a video. \textit{Association} evaluates the capability to search a video pool and retrieve clips that exhibit the same action as the provided query video. \textit{Anticipation} involves predicting future actions based on the video frames observed up to the current timestamp. \textit{Skill Assessment} measures the model's capacity to evaluate the execution proficiency of actions shown in two different videos, classifying which execution is more skillful.

\paragraph{Results.} Tab.~\ref{tab:crossview_bench} summarizes the results. \method\ demonstrates the most significant improvements in \textit{Retrieval} and \textit{Association}, tasks that directly measure semantic equivalence across different viewpoints. Specifically, it achieves absolute gains of $+10.4$ in \textit{Retrieval} and $+11.5$ in \textit{Association} compared to the best baseline, \viewpointrosetta, outperforming both the general-purpose vision-language encoder \sigliptwo\ and existing cross-view methods. This corroborates that the explicit disentanglement of \vi\ and \vv\ features is highly effective for learning action representations robust to viewpoint shifts. Consistent improvements are also observed in \textit{Recognition} ($+7.46$), which measures the transfer of exo-knowledge to ego-views, and \textit{Anticipation} ($+7.32$), which evaluates temporal forecasting capabilities. This confirms the broad applicability of \method's view-invariant representations for both classification and prediction tasks. 

Moreover, on \textit{Skill Assessment}, which evaluates execution proficiency within the same action class, \method\ yields a performance ($55.28$) comparable to \viewpointrosetta\ ($55.82$) and \sigliptwo\ ($55.57$). We attribute this to the fact that proficiency cues (\textit{e.g.,} hand tremors, movement fluidity) rely heavily on subtle visual nuances in execution style rather than the core semantic identity of the action. As a result, such fine-grained visual details are likely allocated to the \vv\ component during the decomposition process.

\subsection{Fine-Grained Temporal Modeling}\label{sec:temporal-modeling}
\paragraph{Evaluation Setup.} We adopt the AE2~\cite{xue2023learning} benchmark to evaluate whether \method\ effectively captures fine-grained temporal dynamics. The benchmark comprises four tasks under two evaluation protocols. Under the zero-shot protocol, models are evaluated directly on frozen features without any task-specific training: \textit{Frame Retrieval} extracts frame-level embeddings from the model's output to evaluate the temporal alignment between the frames of two videos, computed using cosine similarity and reported as the mAP@10 metric; \textit{Phase Ordering} (measured by Kendall's $\tau$) assesses whether the chronological order of events is preserved. Given features from two specific timestamps in a query video, it identifies the two best-aligned timestamps in a different video depicting the same action, and evaluates whether their temporal order is maintained. Under the linear probing protocol, a linear classifier is trained on top of the final layer's frozen features for all models: \textit{Action Phase Classification} performs frame-level action classification, measuring performance via the F1 score; \textit{Phase Progression} applies linear probing to the frame-level features to predict the progression of an action phase on a continuous scale from $0$ to $1$, evaluated using the $R^2$ score. For \textit{Frame Retrieval} and \textit{Action Phase Classification}, we report results under three view settings: \textit{Regular} (the average of intra-view scores, \textit{i.e.,} ego$\to$ego and exo$\to$exo), \textit{Ego2Exo}, and \textit{Exo2Ego}. We compare against both in-domain models trained on AE2 videos and out-of-domain models not exposed to AE2 data.

\begin{figure*}[t]
    \centering
    \includegraphics[width=\linewidth]{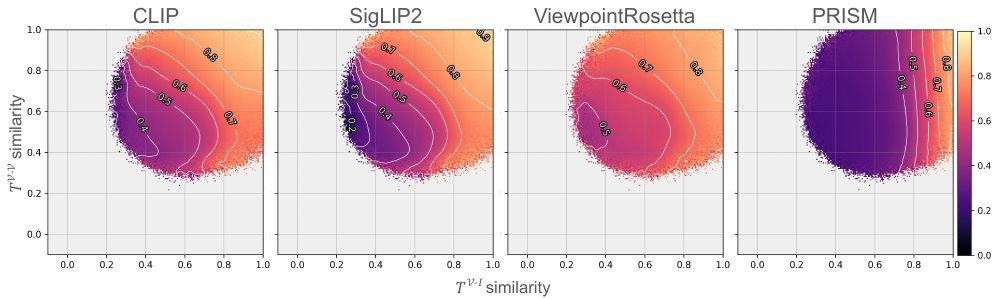}
    % \caption{\textbf{Correlation between representation similarity and textual \vi/\vv\ similarity.} Each point is a clip pair. The $x$- and $y$-axes denote \vv\ and \vi\ textual similarity, respectively, and color indicates representation similarity. For \method, the \vv\ representation is used. Only \method\ shows similarity governed by the \vi\ axis, orthogonal to \vv.}
    % \caption{\textbf{Correlation between representation similarity and textual \vi/\vv\ similarity.} For every clip pair in EgoExoLearn, the $x$- and $y$-axes denote \vv\ and \vi\ textual similarity while color indicates representation similarity (using the \vv\ representation for \method). Only \method\ shows similarity governed by the \vi\ axis, orthogonal to \vv.}
    \vspace{-7mm}
    \caption{\textbf{Correlation between representation similarity and textual \vi\ / \vv\ similarity.} The $x$- and $y$-axes denote \vv\ and \vi\ textual similarity for every clip pair in EgoExoLearn, and color indicates representation similarity (we use $z^\vi$ for \method). Only \method\ shows similarity governed by the \vi\ axis and flat along \vv.}
    \label{fig:correlation}
    \vspace{-2mm}
\end{figure*}
\begin{figure*}[t]
    \centering
    \includegraphics[width=\linewidth]{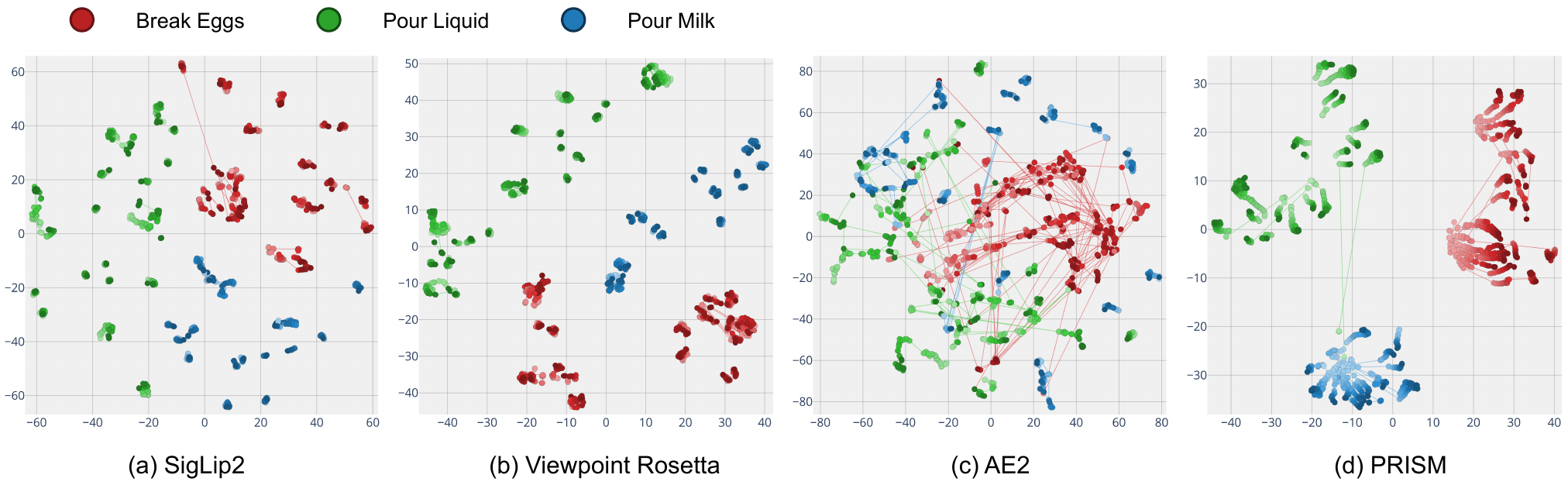}
    \vspace{-7mm}
    \caption{\textbf{t-SNE of frame-level embeddings over time.} Temporally adjacent frames are connected by lines. \method\ is the only model exhibiting both semantic separation and temporal continuity.}
    \label{fig:tsne}
    \vspace{-2mm}
\end{figure*}
\paragraph{Results.} Tab.~\ref{tab:phase_bench} compares \method\ against vision representation baselines and in-domain models trained on AE2 videos. Among out-of-domain models, ours achieves the best performance across all four tasks, outperforming \viewpointrosetta\ by $+16.13$ in \textit{Frame Retrieval}, $+0.55$ in \textit{Phase Ordering}, $+26.64$ in \textit{Action Phase Classification}, and $+0.80$ in \textit{Phase Progression}. More notably, \method\ surpasses the best in-domain model AE2 on \textit{Phase Ordering} ($0.600$ vs.\ $0.562$) and \textit{Phase Progression} ($0.650$ vs.\ $0.480$), while remaining within $3$ and $1$ points on \textit{Frame Retrieval} and \textit{Action Phase Classification}, respectively. This confirms that the self-predictive temporal objective (\S\ref{sec:method-temporal}) internalizes fine-grained temporal dynamics without requiring any in-domain supervision. Additionally, consistent performance is maintained across view settings (\textit{e.g.,} \textit{Frame Retrieval}: $77.00$ regular vs.\ $68.90$/$65.00$ cross-view), demonstrating that \method's temporal representations generalize well across viewpoints.

\begin{figure*}[t]
    \centering
    \includegraphics[width=0.95\linewidth]{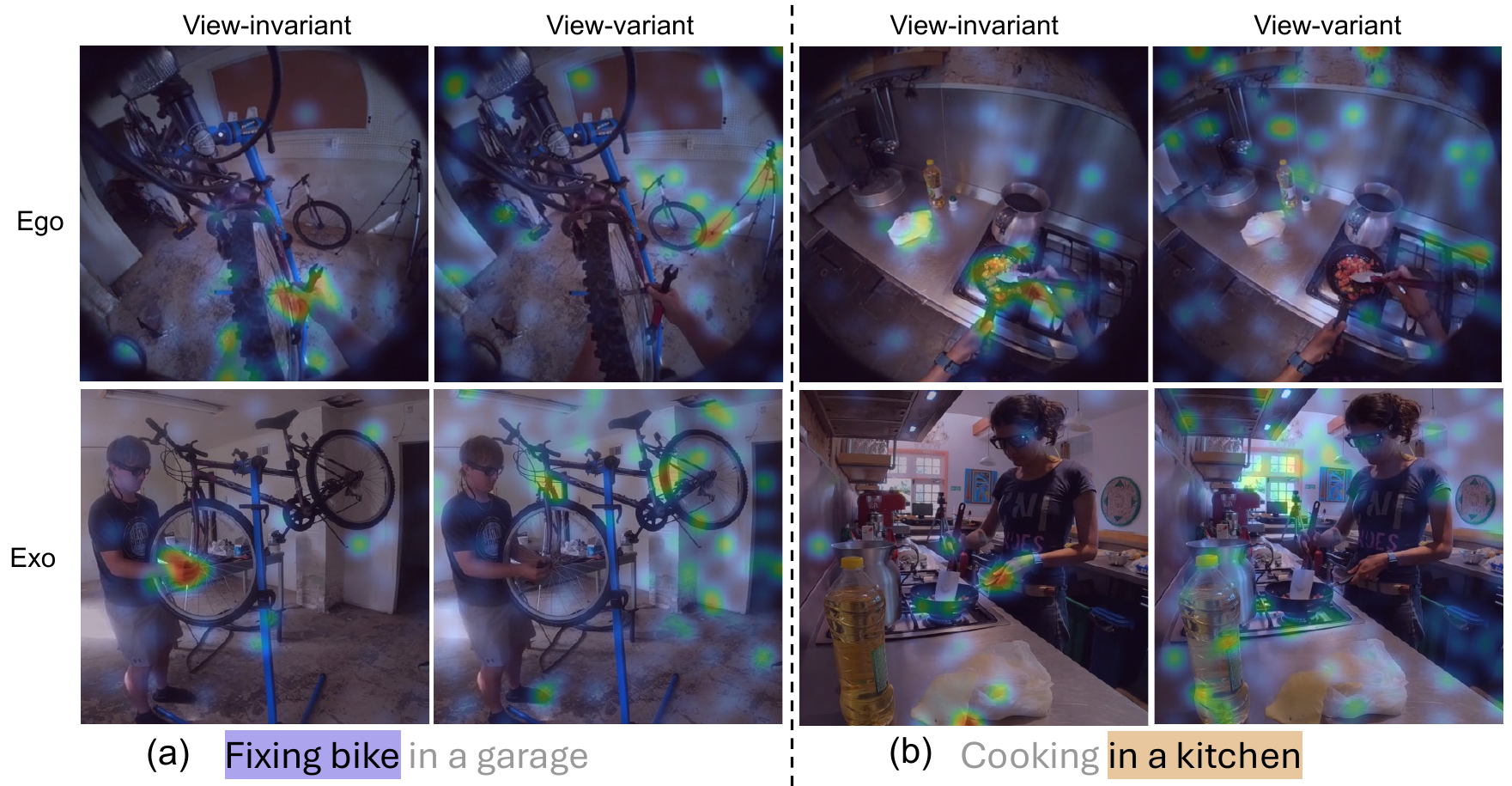}
    \vspace{-3.1mm}
    \caption{\textbf{DeepLift attribution of encoder output patches to the \vi\ and \vv\ streams.} \method\ consistently attends to hands and interacting objects for \vi, and to background regions for \vv, across both ego and exo views.}
    \label{fig:deeplift}
    \vspace{-5mm}
\end{figure*}

\begin{table}[t]
\centering
\small
\setlength{\tabcolsep}{4pt}
% \vspace{-2mm}
\resizebox{0.99\linewidth}{!}{
\begin{tabular}{@{}l cc c cc c@{}}
\Xhline{2\arrayrulewidth}
\multirow{2}{*}[-2pt]{Settings} & \multicolumn{3}{c}{Cross-view Alignment} & \multicolumn{3}{c}{Temporal Alignment} \\
\cmidrule(l{2pt}r{2pt}){2-4} \cmidrule(l{2pt}r{2pt}){5-7}
 & w/o $\mathcal{L}_{\text{temp}}$ & w/ $\mathcal{L}_{\text{temp}}$ & Avg & w/o $\mathcal{L}_{\text{temp}}$ & w/ $\mathcal{L}_{\text{temp}}$ & Avg \\
\hline
Unified      & 38.8 & 32.4 & 35.6 & 64.0 & 69.2 & 66.6 \\
Decomposed   & 50.9 & \textbf{53.5} & \textbf{52.2} & 65.9 & \textbf{72.1} & 69.0 \\
\hline
Avg & 44.9 & 43.0 & -- & 65.0 & \textbf{70.7} & -- \\
\Xhline{2\arrayrulewidth}
\end{tabular}}
\vspace{-1mm}
% \caption{Contribution analysis of training objectives.}
\caption{\textbf{Contribution of training objectives.} \textit{Unified} encodes a video into a single embedding, whereas \textit{Decomposed} splits it into \vi\ and \vv\ streams that are cross-composed across videos. Decomposition drives cross-view alignment while $\mathcal{L}_{\text{temp}}$ drives temporal alignment, and combining both achieves the best on both axes.}
\label{tab:ablation_marginal}
\vspace{-3mm}
\end{table}
\definecolor{headerblue}{HTML}{ECF4FF}
\definecolor{headergray}{HTML}{F2F2F2}
\begin{table}[t]
\centering
\small
\setlength{\tabcolsep}{4pt}
% \caption{Analysis of training-view composition and captioner choice.}
% \caption{Analysis of training views and captioner choice}
% \label{tab:ablation_data}
\vspace{2mm}
\resizebox{1.0\linewidth}{!}{
    \begin{tabular}{@{}l cccccc@{}}
\toprule
\multirow{2}{*}[-2pt]{Settings} & \multicolumn{3}{c}{Cross-view Alignment} & \multicolumn{3}{c}{Temporal Alignment} \\
\cmidrule(l{2pt}r{2pt}){2-4} \cmidrule(l{2pt}r{2pt}){5-7}
 & ego2exo & exo2ego & Avg & Frm. retrieve & Act. phase & Avg \\
\hline
\rowcolor{headergray} \multicolumn{7}{l}{\textit{Supervision w/ Ego-Exo Pairing}} \\
% SUM-L                & 26.2 & 18.4 & 22.3 & 45.8 & 50.9 & 48.4 \\

\actorobservernet    & 20.7 & 18.1 & 19.4 & 44.8 & 34.5 & 39.7 \\
\viewpointrosetta    & 45.8 & 39.2 & 42.5 & 54.2 & 46.9 & 50.6 \\
\hline
\rowcolor{headerblue} \multicolumn{7}{l}{\textit{Supervision w/o Pairing (PRISM)}} \\
Exo only + Gemini    & 52.2 & 44.3 & 48.2 & 69.4 & 73.0 & 71.2 \\
Ego + Exo + Qwen     & 59.4 & 46.3 & 52.9 & 69.2 & \textbf{74.5} & 71.8 \\
\cdashline{1-7}\noalign{\vskip 0.5ex}
Ego + Exo + Gemini   & \textbf{60.1} & \textbf{46.8} & \textbf{53.5} & \textbf{70.5} & 73.6 & \textbf{72.1} \\
\bottomrule
\end{tabular}
}
\vspace{-1mm}
% \caption{Analysis of training views and captioner choice.}
\caption{\textbf{Sensitivity to training views and captioner choice.} \method\ with exo-only data still surpasses all baselines on cross-view alignment, and substituting the captioner introduces only noise-level differences.}
\label{tab:ablation_data}
\vspace{-2mm}
\end{table}

\subsection{Ablation Studies}
\label{sec:ablations}
We analyze how individual components of \method\ contribute to cross-view and temporal alignment. Throughout this section, \textit{Cross-view Alignment} denotes the arithmetic mean of the directional (ego2exo, exo2ego) \textit{Retrieval} (R@$5$) and \textit{Association}$_\text{test}$ scores of Tab.~\ref{tab:crossview_bench}, and \textit{Temporal Alignment} the mean of the \textit{Frame Retrieval} and \textit{Action Phase Classification} scores of Tab.~\ref{tab:phase_bench}.
Tab.~\ref{tab:ablation_marginal} evaluates the effect of explicit \vi/\vv\ decomposition and the self-predictive objective $\mathcal{L}_{\text{temp}}$. The two objectives contribute to largely disjoint axes: decomposition primarily improves cross-view alignment ($35.6 {\to} 52.2$) with minimal impact on temporal alignment, while $\mathcal{L}_{\text{temp}}$ mainly improves temporal alignment ($65.0 {\to} 70.7$) without harming cross-view alignment. Combining both achieves the best performance on both metrics ($53.5$, $72.1$).

Tab.~\ref{tab:ablation_data} analyzes robustness to external dependencies. To simulate the practical setting where paired ego-exo data is unavailable, we remove all ego-view training data (\textit{Exo only}); to evaluate sensitivity to pseudo-caption quality, we replace Gemini with Qwen3VL. Training with only exo-view data moderately reduces cross-view alignment ($53.5 {\to} 48.2$) while largely preserving temporal alignment ($72.1 {\to} 71.2$), indicating that ego-view supervision mainly affects the cross-view axis. Replacing the captioner introduces only marginal differences ($-0.6$ and $-0.3$), suggesting that \method\ is not tightly coupled to a specific captioning model.

\subsection{Analysis on PRISM}\label{sec:analysis}
\paragraph{Correlation of Decomposed Representations.}
To examine whether learned representations correlate with view-invariant or view-variant semantics, we visualize representation similarity against \vi\ and \vv\ textual similarities obtained from a captioning LVLM (Fig.~\ref{fig:correlation}). For \clip, \sigliptwo, and \viewpointrosetta, representation similarity rises with both axes, meaning that different events sharing the same background can still yield high similarity. \method, by contrast, is aligned predominantly with the \vi\ axis and flat along \vv, confirming that its decomposed representation enables event identification independent of background context. This motivates the following quantitative evaluation on counterfactual scenarios.

\paragraph{Robustness to Background Correlation.}\label{sec:background-robustness}
To verify whether the observed semantic separation holds in actual counterfactual scenarios, we evaluate \method\ on the UNSCENE benchmark~\cite{bae2025mash}, which features videos where the action contradicts the background context (\textit{e.g.}, fishing inside a bedroom). Utilizing a subset of $N{=}573$ samples with explicit action captions, we report two metrics: (\lowercase\expandafter{\romannumeral1}) Recall@$10$ (R@$10$), which considers a prediction successful if there is an intersection between the top-$10$ nearest neighbor sets retrieved in the visual representation space and the action caption embedding space, and (\lowercase\expandafter{\romannumeral2}) Representational Similarity Analysis (RSA)~\cite{kriegeskorte2008representational}, which measures the correlation between the two similarity structures. To minimize evaluation bias, we average results across three text encoders (CLIP, SigLIP, Qwen3Embedding).

As shown in Tab.~\ref{tab:model_comparison}, \method\ nearly doubles the best cross-view baseline \viewpointrosetta\ in both metrics ($14.90$ vs.\ $7.50$ in R@$10$; $0.181$ vs.\ $0.098$ in RSA), while matching DINOv2 with only one-tenth of its parameters. This confirms that disentanglement fundamentally requires an explicit decomposition mechanism and cannot be trivially acquired by scaling alone.

\definecolor{headergray}{HTML}{F2F2F2}
\definecolor{rowpurple}{HTML}{D9D2E9}
\begin{table}[t]
\centering
\small
\setlength{\tabcolsep}{2pt}
% \caption{\textbf{UNSCENE Benchmark}: }
% \label{tab:model_comparison}
\resizebox{0.99\linewidth}{!}{
\begin{tabular}{@{}l c cccccc cc @{}}
\Xhline{2\arrayrulewidth}\noalign{\vskip 1pt}
\multirow{2}{*}[-2pt]{Method} & \multirow{2}{*}[-2pt]{\#Params} & \multicolumn{2}{c}{\clip~ViT-L/14} & \multicolumn{2}{c}{\sigliptwo} & \multicolumn{2}{c}{Qwen3Embed} & \multicolumn{2}{c}{Overall} \\
\noalign{\vskip -1pt}\cmidrule(l{2pt}r{2pt}){3-4} \cmidrule(l{2pt}r{2pt}){5-6} \cmidrule(l{2pt}r{2pt}){7-8} \cmidrule(l{2pt}r{2pt}){9-10}
 & & R@\!10 & RSA & R@\!10 & RSA & R@\!10 & RSA & R@\!10 & RSA \\
\hline
% \noalign{\vskip 1pt}
\rowcolor{headerblue} \multicolumn{10}{l}{\textit{Vision Foundation Models}} \\
\vjepatwo         & 1\,B & 7.1 & 0.055 & 7.7 & 0.046 & 7.0 & 0.046 & 7.3 & 0.049 \\
\dinovtwo         & 1\,B & \underline{13.8} & \textbf{0.121} & \underline{14.2} & \underline{0.107} & \underline{14.2} & \underline{0.209} & \underline{14.1} & \underline{0.146} \\
\hline
\rowcolor{headerblue} \multicolumn{10}{l}{\textit{Cross-View Methods}} \\
\actorobservernet & 58\,M  & 6.6 & 0.066 & 6.9 & 0.065 & 6.7 & 0.091& 6.7 & 0.074 \\
% \viencoder      &      & & & & & & & & \\
\suml             & 126\,M & 4.0 & 0.039 & 3.9 & 0.029 & 3.9 & 0.056 & 3.9 & 0.041 \\
\viewpointrosetta & 177\,M & 7.5 & 0.039 & 7.7 & 0.064 & 7.3 & 0.191 & 7.5 & 0.098 \\
\cdashline{1-10}[6pt/4pt] \noalign{\vskip 3pt}
PRISM             & 108\,M & \textbf{14.7} & \underline{0.114} & \textbf{14.9} & \textbf{0.128} & \textbf{15.1} & \textbf{0.301} & \textbf{14.9} & \textbf{0.181} \\
\Xhline{2\arrayrulewidth}
\end{tabular}
}
\vspace{-1mm}
% \caption{\textbf{Robustness to action-scene disentanglement on the UNSCENE benchmark.} We evaluate on videos where actions contradict their background context. \method\ surpasses both vision foundation models and cross-view methods, achieving new SoTA with roughly one-tenth the parameters of DINOv2.}
% \caption{\textbf{Robustness to action-scene disentanglement on UNSCENE.} \method\ surpasses both vision foundation models and cross-view methods with one-tenth the parameters of DINOv2.}
\caption{\textbf{Robustness to action-scene disentanglement on the UNSCENE benchmark.} \method\ surpasses both vision foundation models and cross-view methods.}
\label{tab:model_comparison}
\vspace{-1mm}
\end{table}

\paragraph{Analysis on Semantic-Temporal Dynamics.}
Fig.~\ref{fig:tsne} visualizes t-SNE projections of frame-level embeddings over time for each method. We sample $20$ videos per class from the AE2 benchmark and project their frame-level features into 2D. \sigliptwo\ and \viewpointrosetta\ form video-level clusters but fail to capture temporal progression, while \aetwo\ fails at temporal modeling. In contrast, \method\ produces trajectories that extend continuously over time, confirming that \method\ effectively models fine-grained temporal dynamics.

\paragraph{Visual Attribution of Decomposed Latents.}
To investigate which visual regions drive each decomposed latent, we apply DeepLift~\citep{shrikumar2017learning} to measure the contribution of each image-encoder output patch to the \vi\ and \vv\ streams (Fig.~\ref{fig:deeplift}). In both ego and exo views, the \vi\ stream consistently activates on the actor's hands and interacting objects. Notably, the same semantic targets are highlighted even though they appear at different spatial locations across views, confirming that \vi\ operates on view-invariant cues. The \vv\ stream, in contrast, focuses on the overall background that inherently varies with viewpoint. This spatial separation between the two streams qualitatively demonstrates that \method\ successfully disentangles action-centric cues from peripheral scene context.

\section{Conclusion}

We propose \method\, a framework that decomposes video into view-invariant and view-variant latent streams and recomposes them under language-level supervision, enforcing clean semantic disentanglement robust to counterfactual action--scene compositions. A self-predictive temporal objective operating on an independent axis further internalizes fine-grained temporal dynamics without compromising decomposition quality. Extensive experiments across major cross-view understanding benchmarks demonstrate consistent state-of-the-art performance, validating compositional latent decomposition as a principled and effective approach to view-invariant video representation learning.

% We propose \method\, a framework that decomposes video into view-invariant and view-variant latent streams and recomposes them under language-level supervision to enforce clean semantic disentanglement. Through extensive experiments, we demonstrate that action semantics are effectively captured in the \vi\ representation while leakage of view-variant information is suppressed, yielding robust performance even under counterfactual action--scene compositions. By combining a self-predictive temporal objective that operates on an independent axis, \method\ internalizes fine-grained temporal dynamics without compromising its decomposition capability. As a result, \method\ achieves new state-of-the-art performance across major cross-view understanding benchmarks, including EgoExo4D, EgoExoLearn, and AE2. Our decomposition principle extends naturally to settings where actions must be recognized independently of their visual context, such as robot imitation learning from third-person demonstrations and egocentric skill acquisition in augmented reality.

%\clearpage
%\newpage
\section*{Limitations}
While \method\ demonstrates strong performance in view-invariant representation learning, we acknowledge a few limitations.

First, our semantic decomposition is inherently upper-bounded by the quality of the pre-trained LVLM captioner that provides the language-level supervisory signal. Although Tab.~\ref{tab:ablation_data} shows that swapping the captioner introduces only noise-level differences, this robustness holds only across individual model choices. If a systematic bias is shared across LVLMs, for instance, a tendency to conflate action and scene descriptions due to their co-occurrence in pre-training corpora, such bias would propagate directly into the decomposed supervision targets and, consequently, impose a quality ceiling on the encoder's disentanglement. Addressing this would require either debiased captioning models or additional supervision signals that do not rely on language generation.

Second, all benchmarks evaluated in this work center on procedural human activities involving physical object manipulation (\textit{e.g.,} cooking, sports). This leaves two axes of generalization unexplored: (\lowercase\expandafter{\romannumeral1}) non-human activities, such as animal behavior or natural phenomena, where the notion of view-invariant ``action'' may differ fundamentally from human-centric definitions; and (\lowercase\expandafter{\romannumeral2}) human interactions that lack tangible physical manipulation, such as conversational turn-taking, social gestures, or emotional exchanges, where the relevant semantics may not be neatly separable into action versus scene. Extending the decomposition framework to these broader activity domains constitutes a promising direction for future work.
\section*{Acknowledgments}

This work was supported in part by IITP grant funded by the Korea government (MSIT) (No. RS-2020-II200004, Development of Previsional Intelligence based on Long-Term Visual Memory Network), the Institute of Information \& Communications Technology Planning \& Evaluation (IITP) grant funded by the Korea government (MSIT) (No. RS-2022-II220124), and in part by the KOrea Industrial Technology Association(KOITA) grant funded by the Korea government (No. 2026-KOITA-CO-T2-02-03, Cooperative and Convergent Science and Technology Commercialization Promotion Support Project).

% Bibliography entries for the entire Anthology, followed by custom entries
%\bibliography{anthology,custom}
% Custom bibliography entries only
\bibliography{ref}

\clearpage

\appendix
\clearpage
% \setcounter{page}{1}
% \maketitlesupplementary

% \begin{center}
%   {\Large\textbf{Supplementary Material}}
% \end{center}

\section*{Appendix Contents}

\startcontents[appendices]
\printcontents[appendices]{}{1}{}

\newpage

%\section{Mathematical Justification for PRISM}

% \section{PRISM Algorithm}

\section{Implementation Details}\label{app:impl}

\paragraph{Model Architecture.}
The Decompositional Encoder $\theta$ is built on top of a frozen \sigliptwocite\ (so400m-patch14-384) vision backbone and a frozen Qwen3-Embedding-0.6B~\cite{zhang2025qwen3} text backbone. Per-frame patch embeddings from the vision backbone are fed into a Q-Former of depth $4$ with $8$ attention heads, which produces two per-frame latent vectors of dimension $d_z{=}512$, corresponding to the \vi\ and \vv\ streams. These frame-level latents are then processed by a causal temporal transformer of depth $12$ with $8$ heads, followed by a cross-view transformer of depth $4$ that implements the Compositional Latent Predictor $\phi$. The total number of trainable parameters is approximately $108$M; both the vision and text backbones remain frozen throughout training.

\paragraph{Training Configuration.}
We train \method\ for $6$ epochs on the EgoExo4D~\cite{grauman2024ego} VRS training split using $7$ NVIDIA A6000 $48$GB GPUs, with the remaining GPU reserved for serving the text composer via vLLM. The per-device batch size is $4$ with a gradient accumulation of $2$ steps, yielding an effective batch size of $56$. We use AdamW with a learning rate of $7{\times}10^{-5}$, weight decay of $0.01$, and a \texttt{constant\_with\_warmup} schedule where the warmup phase occupies $10$\% of total training steps. Training is conducted in \texttt{bf16} mixed precision. The loss weights are set to $\lambda_{\text{cross}}{=}1.0$ for $\mathcal{L}_\text{decomp}$ and $\lambda_{\text{next}}{=}0.5$ for $\mathcal{L}_\text{temp}$. The EMA target encoder $\bar{\theta}$ uses a decay coefficient of $\alpha{=}0.998$. Checkpoints are saved every $500$ optimizer steps with a rolling limit of $10$. We use a fixed random seed of $42$ across all experiments.

\paragraph{Video Preprocessing.}
Each video clip is sampled at $4.0$ FPS with a maximum clip duration of $32.0$ seconds, producing up to $T_{\max}{=}128$ frames per sample. Frames are resized to $384{\times}384$ pixels to match the \sigliptwo\ input resolution. Text inputs are tokenized with a maximum sequence length of $128$ tokens.

\paragraph{Cross-view Text Composition.}
As described in \S\ref{sec:method-decompose}, the $\oplus$ operator that fuses $T^\vi_\mathsf{A}$ and $T^\vv_\mathsf{B}$ into a single natural sentence is realized by Qwen3-1.7B served via vLLM. Per-pair composed sentences are cached on disk so that repeated epochs incur only a single LLM call per unique pair. The composed sentence is then encoded by $\mathcal{E}$~\cite{zhang2025qwen3} to produce the target embedding $e_{\mathsf{A},\mathsf{B}}$ used in $\mathcal{L}_\text{decomp}$.

\subsection{Benchmarks}\label{app:benchmarks}

\paragraph{EgoExo4D~\cite{grauman2024ego}.}
EgoExo4D is a large-scale multi-modal, multi-view video dataset comprising $1{,}286$ hours of video across $5{,}035$ takes, captured by $740$ participants in $13$ cities worldwide. Each take simultaneously records egocentric video via Aria glasses and exocentric video from $4$ to $5$ stationary GoPros, all temporally synchronized. The dataset focuses on skilled human activities such as cooking, sports, music, dance, and bike repair, and provides rich annotations including time-indexed natural language descriptions (expert commentary, narrate-and-act, and atomic action descriptions), $3$D body and hand pose, object segmentation masks, keystep labels, and proficiency ratings. Its benchmark suite spans four task families: recognition, proficiency estimation, ego-exo relation, and ego pose.

\paragraph{EgoExoLearn~\cite{huang2024egoexolearn}.}
EgoExoLearn is a dataset of procedural activity videos captured from both egocentric and exocentric viewpoints in real-world environments. In contrast to EgoExo4D, the ego and exo videos are collected asynchronously, \textit{i.e.,} they are not temporally paired, requiring models to establish cross-view correspondence purely through semantic understanding. The dataset is annotated with fine-grained narrations and supports evaluation tasks including cross-view association, action anticipation, and skill assessment.

\paragraph{\aetwo~\cite{xue2023learning}.}
The \aetwo\ benchmark targets fine-grained, frame-level temporal understanding across ego-exo viewpoints. It assembles four action-specific sub-datasets from publicly available sources: \textit{Break Eggs} (CMU-MMAC), \textit{Pour Milk} (H2O), \textit{Pour Liquid} (EPIC-Kitchens and HMDB51), and \textit{Tennis Forehand} (Penn Action and self-collected ego videos). All videos carry dense per-frame action phase annotations, enabling evaluation of temporal alignment quality, phase ordering consistency, frame-level phase classification, and continuous phase progression prediction. The benchmark supports both zero-shot evaluation on frozen features and linear probing protocols, and reports results under intra-view and cross-view settings.

\paragraph{UNSCENE~\cite{bae2025mash}.}
The UNSCENE benchmark, introduced as part of MASH-VLM~\cite{bae2025mash}, is designed to diagnose spurious action-scene correlations, the failure mode first identified by DEVIAS~\cite{bae2024devias}, where models exploit co-occurring background cues rather than action semantics. UNSCENE consists of web-sourced videos depicting counterfactual action-scene compositions, in which the performed action contradicts the typical background context (\textit{e.g.,} fishing inside a bedroom). A subset of $N{=}573$ samples is accompanied by explicit action captions, allowing quantitative evaluation of whether a model's learned representations reflect genuine action identity independently of background context.

% \section{Captioning via LVLM}
\subsection{Captioning via LVLM}
\label{sec:captioning_via_lvlm}

As described in \S\ref{sec:method-decompose}, the language-supervised decomposition objective requires two disjoint textual descriptions per video segment: a \vi\ description $T^\vi$ capturing the agent's action (verbs, hands, tools, target objects, and their spatial relations) and a \vv\ description $T^\vv$ capturing the filming context (camera viewpoint, scene type, background objects, lighting).
$T^\vi$ must read identically regardless of whether the clip is filmed from an egocentric or exocentric viewpoint, while $T^\vv$ must not contain any action verbs or name the tools central to the action.
Given two independently sampled videos $\mathsf{A}$ and $\mathsf{B}$, the text embedding model $\mathcal{E}$~\cite{zhang2025qwen3} maps the recombined text $T^\vi_{\mathsf{A}} \oplus T^\vv_{\mathsf{B}}$ into the target semantic embedding $e_{\mathsf{A},\mathsf{B}}$ that supervises the compositional latent $s_{\mathsf{A},\mathsf{B}}$.
We describe below how $T^\vi$ and $T^\vv$ are generated and how the cross-view composition $\oplus$ is realized.

\paragraph{Captioner model and frame extraction.}
We employ Qwen3-VL-30B-A3B-Thinking~\cite{bai2025qwen3} as the captioning backbone, served via vLLM with one worker per GPU.
For each segment, we extract
$n = \text{clamp}(n_\text{min},\; n_\text{max},\; \lceil d \cdot r \rceil)$
frames via PyAV with keyframe-based seek and decode-time downscaling,
where $d$ is the padded segment duration (${\pm}0.5$\,s), $r{=}2$\,fps is the target sampling rate, $n_\text{min}{=}8$, and $n_\text{max}{=}16$.
Frames are resized to $448 {\times} 448$ at decode time and passed as a $(T, H, W, 3)$ uint8 tensor.

\paragraph{Prompt design.}
Each segment is captioned via a chat-style prompt that instructs the LVLM to produce a JSON object with exactly two keys: $T^\vi$ (\texttt{action\_caption}) and $T^\vv$ (\texttt{context\_caption}), each constrained to ${\le}40$ words.
The user instruction contains three key components:

\textbf{(1) Orientation reasoning block.}
VLMs default to screen-relative left/right, so an exocentric clip filmed facing the agent mirrors left and right relative to the agent's anatomy.
We prepend a step-by-step orientation reasoning protocol to every prompt.
The protocol instructs the model to first locate body landmarks (head, arms, torso), then classify the agent's pose into one of five canonical patterns (egocentric, across-table, frontal facing, back-to-camera, or overhead), each with a deterministic screen-to-anatomy mapping rule.
When the pattern cannot be reliably identified, the model uses side-neutral fallbacks (e.g., ``one hand,'' ``both hands'').
The prompt does not inform the model whether a given clip is ego or exo, so that the model cannot bypass the orientation check.

\textbf{(2) Narration grounding hint.}
When a ground-truth narration is available, it is spliced into the prompt as a grounding hint.
The model is instructed not to paraphrase the hint and to anchor every claim to visual evidence in the frames.

\textbf{(3) Disjointness constraints.}
Explicit negative constraints enforce the structural separation between $T^\vi$ and $T^\vv$: $T^\vi$ must not mention camera, viewpoint, scene type, background, or lighting, while $T^\vv$ must not contain action verbs or name the tool/target pair central to the action.

\paragraph{Output parsing.}
Qwen3-VL-Thinking~\cite{bai2025qwen3} produces a \texttt{<think>...</think>} chain-of-thought block before its final JSON answer.
Our parser strips this block, removes optional markdown fences, regex-extracts the first JSON object, and validates the required fields.
Parse failures are recorded per-record and excluded from training.

\paragraph{Cross-view text composition ($\oplus$).}
\label{sec:llm_composer}
The $\oplus$ operator in $T^\vi_{\mathsf{A}} \oplus T^\vv_{\mathsf{B}}$ is realized by an LLM composer that fuses $T^\vi$ from video $\mathsf{A}$ with $T^\vv$ from video $\mathsf{B}$ into a single natural sentence describing what a clip would look like if it showed the action of $\mathsf{A}$ filmed in the context of $\mathsf{B}$.
We use Qwen3-1.7B served via vLLM for this purpose.
The composer prompt instructs the model to preserve every concrete action detail from $T^\vi_{\mathsf{A}}$ verbatim in meaning while using $T^\vv_{\mathsf{B}}$ only as scene framing, and to output a single fused sentence.
Per-pair results are cached on disk so that repeated epochs incur only one LLM call per unique pair.
The composed sentence is then mapped by $\mathcal{E}$~\cite{zhang2025qwen3} to produce the target embedding $e_{\mathsf{A},\mathsf{B}}$ used in $\mathcal{L}_\text{decomp}$.

% \section{Example Appendix}
% \label{sec:appendix}

% This is an appendix.

\end{document}